\documentclass[letterpaper, 10 pt, conference]{ieeeconf}  

\IEEEoverridecommandlockouts                              

\usepackage{multirow} 
\usepackage{graphicx} 
\usepackage{layouts} 
\usepackage{bbding}
\usepackage{booktabs}
\usepackage{etoolbox}
\usepackage{amsfonts}

\usepackage{amsmath} 
\usepackage[ruled,linesnumbered]{algorithm2e}
\usepackage{comment}
\SetKwRepeat{Do}{do}{while}%

\usepackage{hyperref}
\usepackage[dvipsnames]{xcolor}
\definecolor{myblue}{RGB}{0,0,238}
\hypersetup{
  colorlinks=True,
  linkcolor=black,
  urlcolor=myblue
}

\usepackage{amssymb}
\usepackage{pifont}
\newcommand{\cmark}{\ding{51}}%
\newcommand{\xmark}{\ding{55}}%

\usepackage{cite}

\makeatletter
\patchcmd{\@makecaption}
  {\scshape}
  {}
  {}
  {}
\makeatletter
\patchcmd{\@makecaption}
  {\\}
  {:\ }
  {}
  {}
\makeatother

\title{\LARGE \bf
CFVS: Coarse-to-Fine Visual Servoing for 6-DoF Object-Agnostic Peg-In-Hole Assembly
}

\author{Bo-Siang Lu$^{1}$, Tung-I Chen$^{1}$, Hsin-Ying Lee$^{1}$, and Winston H. Hsu$^{1, 2}$ 
\thanks{$^{1}$National Taiwan University, $^{2}$Mobile Drive Technology}
\thanks{The link of demo video: \url{https://youtu.be/z7DGXmFY4Tk}}
}

\begin{document}

\maketitle
\thispagestyle{empty}
\pagestyle{empty}


\begin{abstract}
%
%
Robotic peg-in-hole assembly remains a challenging task due to its high accuracy demand.
Previous work tends to simplify the problem by restricting the degree of freedom of the end-effector, or limiting the distance between the target and the initial pose position, which prevents them from being deployed in real-world manufacturing. 
Thus, we present a Coarse-to-Fine Visual Servoing (CFVS) peg-in-hole method, achieving 6-DoF end-effector motion control based on 3D visual feedback. CFVS can handle arbitrary tilt angles and large initial alignment errors through a fast pose estimation before refinement.
Furthermore, by introducing a confidence map to ignore the irrelevant contour of objects, CFVS is robust against noise and can deal with various targets beyond training data. Extensive experiments show CFVS outperforms state-of-the-art methods and obtains 100\%, 91\%, and 82\% average success rates in 3-DoF, 4-DoF, and 6-DoF peg-in-hole, respectively.
\end{abstract}

\section{Introduction}
With the trend of industrial automation, intelligent robotic manipulation systems are expected to replace manual work and facilitate precision manufacturing.
Despite the rapid advances in this area, peg-in-hole assembly remains a challenging task due to its low tolerance for deviations: a slight error on keypoint estimation could cause assembly failures.
Though the task has been extensively studied in previous literature, developing a peg-in-hole approach with high accuracy and dexterity toward real-world manufacturing remains an open problem.
Prior work tends to simplify the problem by considering only limited degrees of freedom (DoF) (such as  3-DoF~\cite{luo2018deep,triyonoputro2019quickly,lee2019making,haugaard2020fast} and 4-DoF~\cite{puang2020kovis, valassakis2021coarse}) or poses strict distance constraints between the targets and end-effector~\cite{johannsmeier2019framework,zou2020learning,jin2021contact}.
However, under such simplified settings, these methods could completely fail when dealing with more complicated pose differences between the targets and the end-effector, such as rotations and large initial alignment errors (see Tab.~\ref{table:result}).
Specifically, prior methods can hardly handle rotational deviations, even with just a slight tilt angle, for the insertion direction is assumed to be constantly aligned with the z-axis of the hole during training, 
Moreover, since the methods based on force-torque feedback control need physical contact, and the visual servoing methods suffer from error accumulation, none of the existing methods can effectively handle large initial alignment errors, and thus could fail when the end-effector is far from the targets.

\begin{figure}[t]
  \centering
  \includegraphics[width=0.48\textwidth]{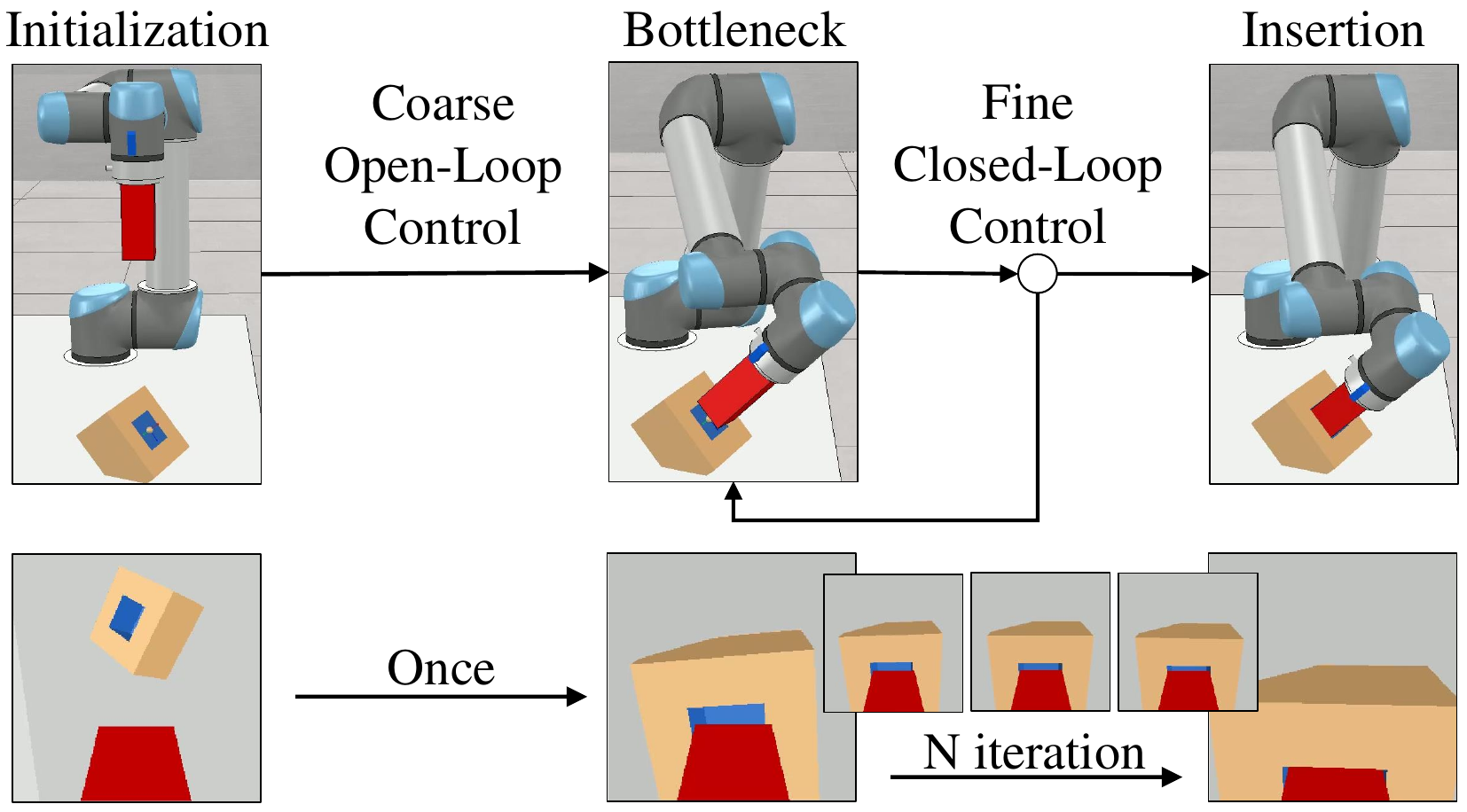}
  \caption{{\bf An overview of CFVS for 6-DoF peg-in-hole.} Integrating the concepts of open-loop control and closed-loop control, we propose CFVS which consists of the coarse and fine approaches. First, we directly move the end-effector to the bottleneck pose, overcoming the challenge of the large initial error. Then, we align the peg to the hole gradually with visual feedback, achieving a high insertion success rate in 6-DoF peg-in-hole.}
  \label{fig:overview}
\end{figure}


%
To address these issues, we present a novel Coarse-to-Fine Visual Servoing (CFVS) 6-DoF peg-in-hole approach that can tackle large initial alignment errors and perform 6-DoF motion control by leveraging 1) 3D point-cloud information and 2) a coarse-to-fine offset prediction pipeline. The proposed CVFS consists of two major components, the open-loop control with object-agnostic keypoint network (OAKN) and the visual servoing with offsets prediction network (OPN). The OAKN estimates the rough target position and guides the end-effector towards the objective. The OPN then gradually aligns the peg to the hole according to the received 3D visual feedback. In comparison to pure closed-loop control methods, this coarse-to-fine strategy alleviates large initial alignment errors by rapidly decreasing the range of exploration, avoiding error accumulation caused by step-by-step refinement. 
Instead of taking 2D images \cite{triyonoputro2019quickly,haugaard2020fast,puang2020kovis}, CFVS operates with 3D point-cloud inputs, which offer depth information of input objects and enable 6-DoF movement control. CFVS can therefore achieve 6-DoF insertion by encoding the 3D spatial relationship between pegs and targets. In addition, by adopting a confidence map that concentrates on the information of holes and disregards that of object contours, CVFS captures the local perception of target objects, which makes the insertion robust and agnostic to the variation of object shapes.

We outperform all state-of-the-art approaches and obtain 100\%, 91\%, and 82\% average success rates in 3-DoF, 4-DoF, and 6-DoF peg-in-hole with 30 cm initial error. Experiment results demonstrate that CFVS succeeds in tackling large initial errors and diverse shape variations. Ablations (Sec. \ref{sec: ablations}) also highlight the importance of our coarse-to-fine strategy.

In brief, our main contributions are as follows:
\begin{itemize}
    \item We propose CFVS, a coarse-to-fine 3D point-based visual servoing framework, which is the first to achieve 6-DoF peg-in-hole assembly with tilted holes.
    \item CFVS can achieve accurate insertion even with large initial alignment errors. 
    \item CFVS is robust to various shapes of target objects and can generalize to unseen objects.
\end{itemize}

\section{Related Work}
\subsection{Peg-In-Hole Assembly Task}
Many researchers focus on the peg-in-hole problem which has no general formulation to solve \cite{xu2019compare}. Some works \cite{fei2003assembly,kim2014hole,tang2016autonomous,tang2016teach,inoue2017deep,luo2018deep,van2018comparative,johannsmeier2019framework,liu2019screw,zou2020learning,beltran2020variable,jin2021contact} regard peg-in-hole assembly as a contact-rich problem. They use the force-torque sensor to search a hole so that they can achieve high accuracy. However, in their method, the end-effector needs physical contact with target objects.

There exist some approaches using the visual sensor, as shown in Table \ref{table:related_paper}. For instance, M. Nigro et al. \cite{nigro2020peg} leverage neural networks and traditional algorithms to detect the hole position and insertion direction. Although they can insert the peg into the tilted hole, they only use the open-loop method without further refinement. Compared to the method, the closed-loop approaches are more robust. J. C. Triyonoputro et al. \cite{triyonoputro2019quickly} and R. L. Haugaard et al. \cite{haugaard2020fast} propose a learning-based visual servoing method to achieve 3-DoF peg-in-hole. Some methods \cite{lee2020guided,bogunowicz2020sim2real} use RL policy to accomplish the 3-DoF task. For 4-DoF insertion task, E. Valassakiand et al. \cite{valassakis2021coarse} insert a square or cylindrical ring over a peg. They use Iterative Closest Point (ICP) \cite{besl1992method} to approximately achieve a  bottleneck pose followed by a closed-loop policy. E. Y. Puang et al. \cite{puang2020kovis} use an autoencoder, which reconstructs depth images from grayscale images, to learn self-supervised keypoint features in latent space. Then, train a visual servo network with keypoint features for insertion. Existing methods can not learn the 3D spatial relationship between the peg and hole and control the end-effector in 6-DoF. On the contrary, we utilize 3D information and are the first to achieve 6-DoF peg-in-hole with large initial alignment errors.

\subsection{Coarse-to-Fine Strategy for Robotic Manipulation}
The concept of coarse-to-fine strategy has been conducted for a long time \cite{salcudean1989control,sharon1984enhancement}, which is a common method for robotic manipulation applications. Such a strategy can boost the performance and thus achieve a high success rate. E. Johns \cite{johns2021coarse} uses an imitation learning method to learn for robotic manipulation. Their framework can be split into two phases: approach and interaction trajectory. E. Valassakis et al. \cite{valassakis2022demonstrate} introduce a one-shot imitation learning method that learns from a demonstration. They move to a bottleneck pose based on visual servoing and then replay the demonstration. M. A. Lee et al. \cite{lee2020guided} use both model-based and model-free approaches. They estimate a rough pose to approach the object and then adopt a model-free policy for refinement. E. Valassaki et al. \cite{valassakis2021coarse} propose a learning-based control policy. At first, they use Iterative Closest Point (ICP) \cite{besl1992method} to estimate the object pose and then refine the pose to increase accuracy. Unlike these methods, which need to record the demonstration for each object in prior or can only operate with the same object, our coarse-to-fine pipeline can be agnostic to objects.

\begin{table}[t]
\vspace{6pt}
\caption{Comparison with vision-based approaches.}
\label{table:related_paper}
\begin{center}
\setlength{\tabcolsep}{1mm}
\vspace{-4mm}
\begin{tabular}{lcccc}
\hline
\multirow{2}{*}{Method} & 6-DoF & Unseen & Large Initial & Closed-loop \\
& Peg-in-hole & Objects & Alignment Error & Control\\
\hline
\cite{triyonoputro2019quickly,haugaard2020fast,puang2020kovis} & \xmark & \xmark & \xmark & \cmark\\
\cite{lee2020guided,valassakis2021coarse,bogunowicz2020sim2real} & \xmark & \xmark & \cmark & \cmark \\
\cite{nigro2020peg} & \xmark & \xmark & \cmark& \xmark\\
CFVS (Ours) & \cmark & \cmark & \cmark& \cmark\\
\hline

\end{tabular}
\end{center}
\end{table}
\begin{figure*}[t]
  \centering
  \includegraphics[width=1.0\textwidth]{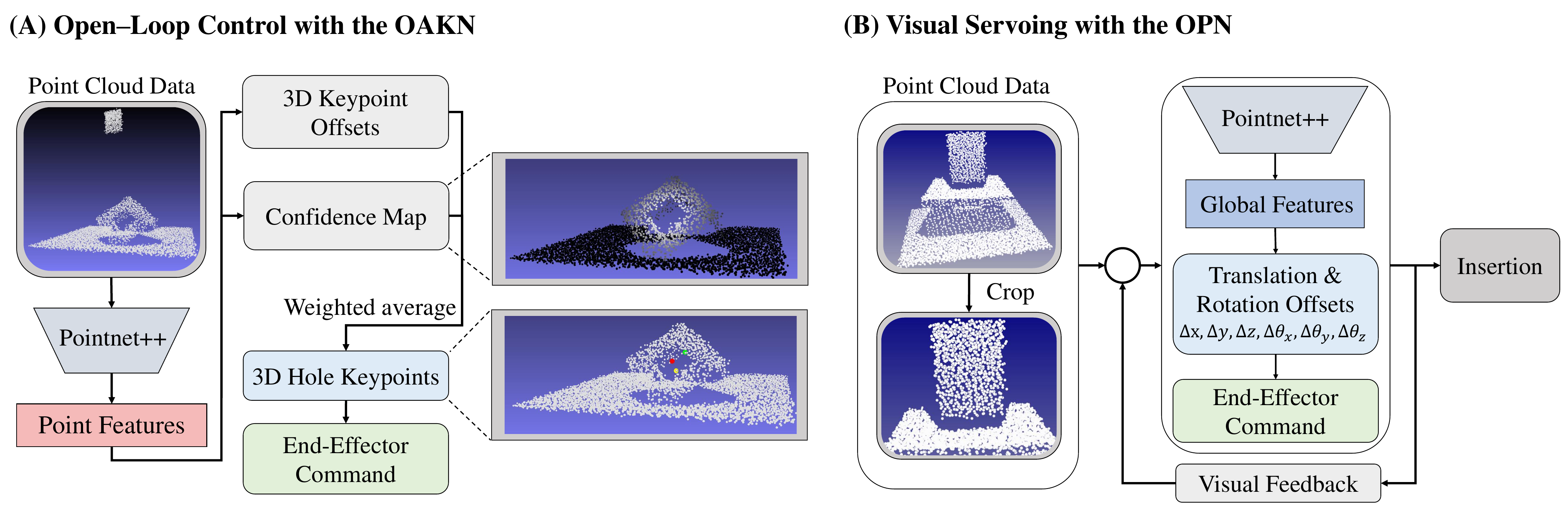}
  \caption{{\bf The 6-DoF peg-in-hole pipeline} contains two stages: (A) the open-loop control with the object-agnostic keypoint network (OAKN), which estimates the rough pose of a hole, and (B) visual servoing with the offset prediction network (OPN), which iteratively refines the final insertion position. In the first stage, OAKN takes a 3D point cloud as input and predicts candidate keypoints from each point. These keypoints, which represent the position and orientation of the hole, are weightedly averaged by the confidence heatmap that highlights points close to the hole and makes our pipeline object-agnostic. The end-effector then moves to that position and orientation. In the second stage, OPN extracts global features from the point cloud cropped around the hole and then generate the translation and rotation offsets for the end-effector according to the features. The OPN is iterated until we reach sufficiently small offsets. The insertion command is finally executed.}
  \label{fig:architecture}
\end{figure*}

\section{Method}

We present CFVS, a coarse-to-fine 6-DoF peg-in-hole method comprised of two components: (A) the open-loop control with an object-agnostic keypoint network (OAKN) and (B) the visual servoing with an offsets prediction network (OPN), as shown in Fig.~\ref{fig:architecture}. 
The input 3D point clouds are reconstructed from depth images captured by the eye-in-hand camera in the world coordinate.
We adopt PointNet++~\cite{qi2017pointnet++} as the backbone network, which is a widely-adopted feature extractor for 3D robotic manipulation, to encode 3D points into high-dimensional features for further processing.
%

\subsection{Object-Agnostic Keypoint Network (OAKN)}
\label{sec:coarse}



%
OAKN is designed to guide the end-effector towards the target position near the hole before further refinement. 
We propose to leverage point cloud features, which contain more abundant geometrical information than 2D images, to predict the required position and orientation information for open-loop control. The details are described as follows.

%
%

\subsubsection{Open-Loop Control}
\label{sec:olc}
Inspired by \cite{he2020pvn3d}, the pose information for open-loop control can be represented by three keypoints, $K = \{k_{1},k_{2},k_{3}\}$.
Specifically, $k_{1}$ denotes the central position $t \in \mathbb{R}^3$ of the hole, and the orientations along x- and z-axis can be computed by $v_x = k_2 - k_1$ and $v_z = k_3 - k_1$, respectively.
The rotation matrix $R \in \mathbb{SO}^3$ can be computed according to these keypoints (see Algorithm~\ref{alg:rot}), and the end-effector can be guided to the target pose $[R|t] \in \mathbb{SE}^3$ through inverse kinematics.
The challenge lies in how to obtain high-quality keypoints based on unstructured 3D points captured by the depth camera.
In the following sections, we will discuss how to generate candidate keypoints based on non-euclidean point cloud data and how to adaptively combine these candidates into the keypoints close to the target position.  
%
%
\subsubsection{Keypoint Prediction}
\label{sec:keypoint_offset}
Let $X = \{x_i\}_{i=1}^{N}$ be the input point cloud, where $x_i$ is a point in 3D coordinates and $N$ is the number of points. 
The backbone network encodes $X$ into the feature $z \in \mathbb{R}^{N \times C}$, which will be sent to a Multi-Layer Perceptron (MLP) to predict the keypoint offsets $\{\Delta k_{i,1}, \Delta k_{i,2}, \Delta k_{i,3}\}$.
Thus, through computing $k_{i,j} = x_i + \Delta k_{i,j}$, the $N$ candidate keypoints $K_i$ can be obtained:
\begin{equation}
K_i = \{k_{i,1},k_{i,2},k_{i,3}\}, \quad i\in \{1, 2, \ldots, N\}.
\end{equation}
Naturally, one intuitive way to obtain the ultimate objective $K$ for open-loop control is to average across all $K_i$. However, our experiment shows that such a naive average-pooling strategy leads to sub-optimal performance (as shown in Table~\ref{table:ablation}). 
Therefore, we propose combining the candidate keypoints with a learnable confidence map that can adaptively re-weight the importance of $K_i$. 

\subsubsection{Confidence Map}
\label{sec:heatmap}
After having multiple candidates, our next step is to determine the final keypoints to guide the end-effector towards the hole. 
However, since the central position of the hole is unknown at inference, we cannot directly assign the confidence score to each candidate keypoint according to the distance. 
Therefore, we introduce a learnable confidence map to re-weight each candidate keypoint and obtain the final keypoints.
The confidence map is also predicted by a MLP, which receives $z$ as the input and outputs $\{w_i\}_{i=1}^N,\, w_i \in [0,1]$.
Thus, the final keypoints $K$ can be obtained by re-weighting each candidate keypoint: 
\begin{equation}
    k_j = \frac{\sum^N_{i=1} w_i k_{i,j}}{\sum^N_{i=1}w_i},\quad j \in \{1,2,3\}.
\end{equation}
The negative influence caused by irrelevant visual features, such as object contours and noisy backgrounds, can be eliminated through such an attention mechanism.
As a result, the proposed model is robust against shape variation and can be applied to multiple different target objects (see Fig.~\ref{fig:dataset}).

\begin{algorithm}
\caption{Computation of rotation matrix}\label{alg:rot}
\DontPrintSemicolon
\KwIn{3D hole keypoints, $k_1$, $k_2$ and $k_3$.}
\KwOut{Rotation matrix, $R$.}
  \SetKwFunction{FMain}{compute\_rot\_mat}
  \SetKwProg{Fn}{Function}{:}{}
  \Fn{\FMain{$k_1$, $k_2$, $k_3$}}{
    $v_x \leftarrow k_2 - k_1$\;
    $v_z \leftarrow k_3 - k_1$\;
    $v_z^{norm} \leftarrow$ normalize\_vector($v_z$)\;
    $v_y \leftarrow$ cross\_product($v_z$, $v_x$)\;
    $v_y^{norm} \leftarrow$ normalize\_vector($v_y$)\;
    $v_x^{norm} \leftarrow$ cross\_product($v_y$, $v_z$)\;
    $R \leftarrow \begin{bmatrix} |&|&| \\ v_x^{norm} & v_y^{norm} & v_z^{norm} \\ |&|&| \end{bmatrix}$\;
    \KwRet $R$\;
  }
\end{algorithm}

\subsubsection{Loss} 
\label{sec:coarse_loss}
%
OAKN is trained to minimize the loss $L_\text{coarse} =  L_\text{kpts} +  L_\text{map}$, where $L_\text{kpts}$ is the loss of the predicted keypoint offsets and $L_\text{map}$ represents the loss of the confidence map.
Specifically, $L_\text{kpts}$ is formulated as the weighted L1 loss between the predicted and ground truth keypoint offsets:
\begin{equation}
        L_\text{kpts} = \frac{1}{N}\sum^N_{i=1}\sum^3_{j=1} w_i^* \Vert \Delta k_{i,j} - \Delta k^{*}_{i,j} \Vert,
\end{equation}
where $w_i^*$ denotes how important the candidate keypoints $K_i$ are, and $\Delta k^{*}_{i,j}$ is the ground-truth 3D keypoint offset. 
Since the importance of $K_i$ does not have ground truth, $w_i^*$ is generated from a 3D Gaussian function:

\begin{equation}
    w_i^* = e^{-\frac{1}{2\sigma^2}(\Vert x_i - p\Vert^2)},
\end{equation}
where $p \in \mathbb{R}^3$ is the central position of the hole and $\sigma$ controls the range of the confidence map. Points closer to the position of the center are assigned higher confidence. This weighting mechanism makes OAKN ignore the contour of objects and backgrounds, paying more attention to the information near the hole center. 
On the other hand, $L_\text{map}$ is formulated as the root mean square error, which forces the predicted confidence map to approximate a Gaussian distribution centered at the hole:
\begin{equation}
    L_\text{map} = \sqrt{\frac{1}{N}\sum^N_{i=1}\Vert w_i - w_i^* \Vert^2}.
\end{equation}




\subsection{Visual Servoing with the Offsets Prediction Network}
\label{sec:fine}

We adopt the visual servoing for further refinement in the second stage. 
OAKN predicts the pose of the end-effector close to the hole, but the pose is not accurate for insertion.
Thus, we refine the pose via visual servoing with OPN, which estimates the translation and rotation offsets iteratively with visual feedback until we reach sufficiently small offsets or repeat up to the specific times. 

\subsubsection{Translation and Rotation Offsets}
\label{sec:oft}
We refine the pose of the end-effector for precise insertion by estimating offsets that represent the pose difference between the current pose and the ideal target pose. We first crop the input point cloud $\{x_i\}_{i=1}^{N}$ into $\{x_i\}_{i=1}^{N'}$, where $N' \in [0,N]$ is the number of points near the end-effector. By cropping the point cloud, we focus more on the peg and hole and ignore other redundant information. Given the cropped point cloud, the OPN extracts the global features $\in \mathbb{R}^{D}$, which are fed into an MLP to predict the translation offsets $\Delta t = (\Delta x, \Delta y, \Delta z) \in \mathbb{R}^{3}$ in 3D coordinate and rotation offsets $\Delta r = (\Delta \theta_x, \Delta \theta_y, \Delta \theta_z) \in \mathbb{R}^{3}$ in Euler angle representation.


\subsubsection{Visual Servoing}
\label{sec:clc}

We use visual servoing, a closed-loop control method, to refine the pose of the end-effector for insertion by iteratively estimating the translation offsets $\Delta t$ and rotation offsets $\Delta r$ with the OPN. We first record the original pose of the end-effector $[R|t] \in \mathbb{SE}^3$. The rotation offset $\Delta r$ is converted to the representation of rotation matrix $\Delta R \in \mathbb{SO}^3$. Then, we move the end-effector to the next pose $[R'|t'] \in \mathbb{SE}^3$, where $R' = \Delta R \cdot R$ and $t' = \Delta t + t$. Our fine approach, outlined in Algorithm \ref{alg:vs}, executes repeatedly until the predicted offsets are smaller than the error tolerance or the OPN is repeated up to specific times. Finally, we execute the insertion command to accomplish the task.

\begin{algorithm}
\caption{Visual servoing}
\label{alg:vs}
\DontPrintSemicolon
\KwIn{Point cloud, $\{x_i\}_{i=1}^{N'}$. Translation error tolerance, $e_t$. Rotation error tolerance, $e_r$.}
\KwOut{End-effector pose, $ee$.}
  \While{True}{
    $\Delta t, \Delta \theta_r \leftarrow OPN(\{x_i\}_{i=1}^{N'})$\;
    $\Delta R \leftarrow convert\_to\_rotation\_matrix(\Delta \theta_r)$\;
    $R, t \leftarrow get\_ee\_pose()$\;
    $R' = \Delta R \cdot R$\;
    $t' = \Delta t + t$\;
    set $ee$ to $R', t'$\;
    \If{$\Delta t < e_t$ and $\Delta \theta_r < e_r$}{
      break \tcp*{servoing done}}
  }
\end{algorithm}

\subsubsection{Loss}
\label{sec:fine_loss}
We train the OPN with supervised learning by minimizing the loss $L_\text{fine} =  L_\text{trans} +  L_\text{rot}$, where $L_\text{trans}$ and $L_\text{rot}$ are the losses of translation and rotation offsets. For the translation offsets, we use both the root mean square error and cosine distance to optimize the translation error:
\begin{equation}
    L_\text{trans} = \sqrt{\frac{1}{N}\sum^N_{i=1}\Vert \Delta t - \Delta t^* \Vert^2} + (1 - \frac{\Delta t^T \Delta t^*}{\vert \Delta t \vert \vert \Delta t^* \vert}),
\end{equation}
where $\Delta t^* = (\Delta x^*,\Delta y^*,\Delta z^*)$ is the ground-truth translation offsets. We use the root mean square error to learn the magnitude of translation offsets and cosine distance to learn the direction of movement. For the rotation offsets, we adopt the root mean square error to minimize the rotation error:
\begin{equation}
    L_\text{rot} = \sqrt{\frac{1}{N}\sum^N_{i=1}\Vert \Delta r - \Delta r^* \Vert^2},
\end{equation}
\noindent where $\Delta r^* = (\Delta \theta_x^*,\Delta \theta_y^*,\Delta \theta_z^*)$ is the ground-truth rotation offsets.

\begin{figure}[t]
\vspace{6pt}
  \centering
  \includegraphics[width=0.48\textwidth]{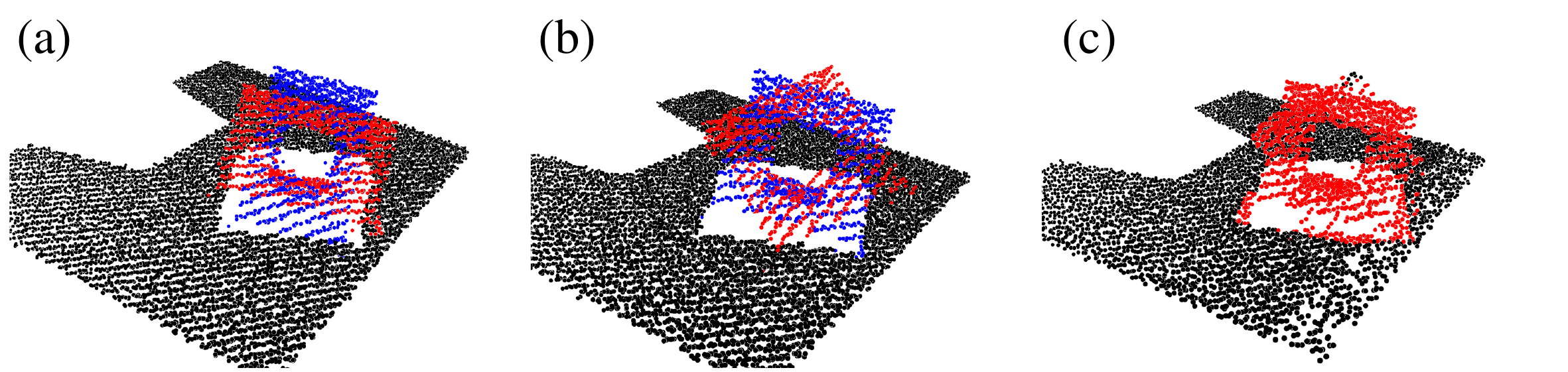}
  \caption{{\bf Data augmentation.} We apply three different augmentation: (a) random scaling, (b) random rotation and (c) mix of these two methods. (Best seen in color.)}
  \label{fig:aug}
\end{figure}
\subsection{Data generation}

We collect coarse dataset $D_\text{coarse}$ and fine dataset $D_\text{fine}$ on Coppeliasim (formerly V-REP) \cite{rohmer2013v}. The coarse dataset is trained on the OAKN while the fine dataset is trained on the OPN. For both the coarse and fine datasets, the hole object is randomly translated and rotated within the workspace.

\subsubsection{Coarse Dataset}
The coarse dataset is denoted as $D_\text{coarse} = \{X_i, \Delta K_i, W_i\}_{i=1}^M$, where M is the total number of the coarse data and $X_i$, $\Delta K_i$ and $W_i$ are the point cloud, 3D keypoint offsets and confidence map under one scene. We collect the coarse dataset with M iterations. For each iteration, we keep the end-effector at an initial configuration and record the coarse data.

\subsubsection{Fine Dataset}
The fine dataset is defined as $D_\text{fine} = \{X_i, A_i\}_{i=1}^L$, where L is the total number of the fine data and $X_i$ and $A_i$ are the point cloud and actions under one scene. We collect the fine dataset with L iterations reversely. For every iteration, we set the pose of the end-effector to the ground-truth hole pose initially. We randomly decide the $-\Delta t$ and $-\Delta \theta_r$, which are the negative value of ground-truth translation and rotation offsets. Then, we move the end-effector according to the $-\Delta t$ and $-\Delta \theta_r$ and record the $\Delta t$ and $\Delta \theta_r$ as fine data.

\subsubsection{Data augmentation}

Our training dataset is augmented with random scaling, random rotation, and a mix of these two methods, as shown in Fig. \ref{fig:aug}. We augment each object only along the x-axis and y-axis of the object frame. The points near the center of the holes are omitted to prevent the deformation of the holes.
The augmentation increases the richness of the training set and boosts performance.

\section{Experiments}
We conduct a series of experiments to measure the performance of CFVS. We are curious about the accuracy, generalization capability, and efficiency. Thus, we want to examine: (1) How does our proposed method compare to other baselines in 3-DoF, 4-DoF, and 6-DoF peg-in-hole assembly with the small and large initial alignment error? (2) Can our proposed method generalize to unseen objects despite the large shape variation? (3) How fast is our overall framework with the small and large initial alignment error?

\subsection{Experimental Settings}
The experiments are running on Coppeliasim \cite{rohmer2013v} with UR5 robot. We deploy our method on such an extendable simulation for a better comparison and will release the code soon. We control the end-effector to the desired pose in 3D coordinate by inverse kinematics automatically solved by the simulator. In the simulated environment, the RGB-D camera is eye-in-hand, and the end-effector is a round or square peg strictly attached to the robotic arm. That is, there is no relative motion between the peg and the robotic arm. When testing, we add Gaussian noise $\mathcal{N}(0~\text{mm},1~\text{mm}^{2})$ on a point cloud converted from a depth image. The radius and height of the round peg are 2.3 cm and 10 cm while the radius and depth of the round hole are 2.5 cm and 4.5 cm. The square peg is a 4.6 cm $\times$ 4.6 cm $\times$ 10 cm cuboid while the square hole is a 5 cm $\times$ 5 cm $\times$ 4.5 cm rectangular cavity. The clearance between the peg and hole is set to 4 mm. We set the initial distance between the peg and hole to 15 cm as the small initial alignment error and 30 cm as the large one.

\begin{figure}[t]
\vspace{6pt}
  \centering
  \includegraphics[width=0.48\textwidth]{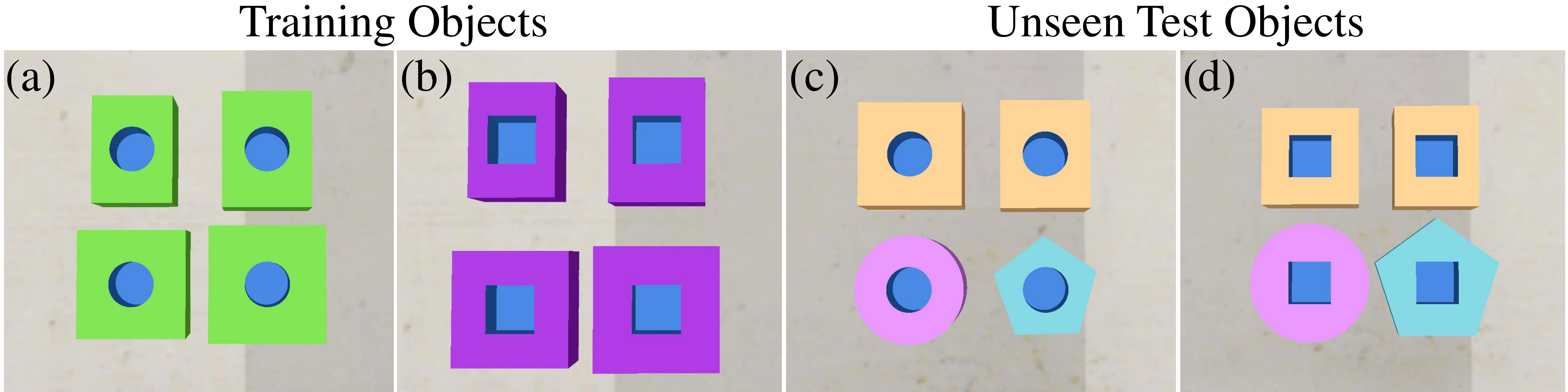}
  \caption{{\bf Train and Unseen Test Objects.} (a)(b) are training objects while (c)(d) are test objects. (a)(c) are round holes which are used for 3-DoF peg-in-hole, while (b)(d) are square holes which are used for 4-DoF and 6-DoF peg-in-hole.}
  \label{fig:dataset}
\end{figure}

\subsection{Tasks}
In our experiments, we design the 3-DoF, 4-DoF, and 6-DoF tasks. We test each object with 15 cm and 30 cm initial alignment error, respectively. All of these three tasks are with 4 mm clearance, the goal of the tasks is to insert a peg into a hole. For the shape of objects, we take four cuboids for training, and then use two similar unseen cuboids with small variation and two unseen shapes with large variation for testing, as shown in Fig. \ref{fig:dataset}.

\smallskip\noindent\textbf{3-DoF Peg-in-hole.} For 3-DoF, the peg and hole are round, and the object is randomly translated in XY-plane within the workspace. This is the simplest task because the end-effector can only move in the position of x, y, and z.

\smallskip\noindent\textbf{4-DoF Peg-in-hole.} For 4-DoF, the peg and hole are square. The object can be randomly translated in XY-plane and rotated along the z-axis, which is perpendicular to the workspace plane. This task is more difficult because the end-effector needs to learn the movement of x, y, z, and yaw.

\smallskip\noindent\textbf{6-DoF Peg-in-hole.} For 6-DoF, it is the most difficult of our tasks, and the peg and hole are square. The object can be randomly translated in XY-plane and rotated along the x-axis, y-axis, and z-axis. We set the tilt angle which is between the vertical axis and the direction of insertion to [0, 50] degrees due to the limitation of the end-effector. If the tilt angle is too large and the hole is back to the robotic arm, the end-effector can not reach such a pose. That is, there is no solution for inverse kinematics. The end-effector needs to learn the movement of x, y, z, roll, pitch, and yaw.

\subsection{Evaluation Metrics}
We use success rate to measure the performance. The depth of all the hole objects is 45mm. After the end-effector executes the insertion, if the peg touches the bottom of the hole, we regard this case as successful. Otherwise, it is a failure. Our proposed approach is compared to three different baselines. We test each unseen object 250 times, and the total is $250\times4 = 1000$ trials for each task. For the visual servoing methods, we additionally experiment with efficiency. 

\subsection{Baselines}

\smallskip\noindent\textbf{ICP \cite{besl1992method}.} This approach, a conventional method for point cloud registration, is not a learning policy. We use the 3D model of 7 mm $\times$ 13 mm $\times$ 13 mm cuboid in the training set to estimate the object pose. The initial transformation matrix is set to the transformation which is from the position of the end-effector to the center point of the workspace.

\smallskip\noindent\textbf{ICP w/ kpts \cite{besl1992method}.} Similar to the ICP, ICP w/ kpts is additionally given a rough object pose. The pose is predicted by neural networks for the initial transformation matrix.

\smallskip\noindent\textbf{3DRHD \cite{nigro2020peg}.} This is an open-loop method estimating the hole position and insertion direction with an RGB-D camera. We implement the paper and use the point cloud which involves the whole scene instead of the object surface. 
\smallskip\noindent\textbf{KOVIS \cite{puang2020kovis}.} We train KOVIS, a 2D visual servoing with the default settings. KOVIS is originally tested for their 4-DoF task with the round peg and hole. In our 4-DoF and 6-DoF tasks, we test KOVIS with the square peg and hole.
\begin{table}[htbp]
\vspace{6pt}
\caption{Success rate of peg-in-hole assembly with 4mm clearance. We test on 3-DoF, 4-DoF and 6-DoF peg-in-hole assembly with 15cm and 30cm initial alignment error. The result shows that ours outperforms to all other baselines with tilted holes and large initial alignment errors.}
\label{table:result}
\begin{center}
\setlength{\tabcolsep}{0.5mm}
\vspace{-6mm}
\begin{tabular}{@{\extracolsep{2pt}}c||c|cccc@{}}

\multicolumn{6}{c}{\multirow{2}{*}{(a) 3-DoF peg-in-hole assembly (15 cm / 30 cm initial alignment error)}} \\
\multicolumn{6}{c}{} \\
\hline
\multirow{3}{*}{Method} & \multirow{3}{*}{Avg.} & \multicolumn{2}{c}{Small shape variation} & \multicolumn{2}{c}{Large shape variation} \\
\cline{3-4} \cline{5-6} 
& & \multirow{2}{*}{$\rm Cuboid_1$} & \multirow{2}{*}{$\rm Cuboid_2$} & \multirow{2}{*}{Cylinder} & Pentagonal \\
&&&&& prism \\
\hline
ICP \cite{besl1992method} & 0.10/0.10 & 0.11/0.10 & 0.10/0.11 & 0.11/0.11 & 0.07/0.08 \\
ICP w/ kpts \cite{besl1992method} & 0.67/0.69 & 0.71/0.74 & 0.77/0.76 & 0.69/0.70 & 0.52/0.54 \\
3DRHD \cite{nigro2020peg} & 0.69/0.68 & 0.80/0.79 & 0.73/0.71 & 0.63/0.62 & 0.61/0.61 \\
KOVIS \cite{puang2020kovis} & 0.95/0.80 & \textbf{1.00}/0.88 & \textbf{1.00}/0.87 & 0.86/0.71 & 0.94/0.75 \\
CFVS (Ours) & \textbf{1.00}/\textbf{1.00} & \textbf{1.00}/\textbf{1.00} & \textbf{1.00}/\textbf{1.00} & \textbf{1.00}/\textbf{1.00} & \textbf{1.00}/\textbf{1.00} \\
\hline
\multicolumn{6}{c}{} \\
\multicolumn{6}{c}{\multirow{2}{*}{(b) 4-DoF peg-in-hole assembly (15 cm / 30 cm initial alignment error)}} \\
\multicolumn{6}{c}{} \\
\hline
\multirow{3}{*}{Method} & \multirow{3}{*}{Avg.} & \multicolumn{2}{c}{Small shape variation} & \multicolumn{2}{c}{Large shape variation} \\
\cline{3-4} \cline{5-6} 
& & \multirow{2}{*}{$\rm Cuboid_1$} & \multirow{2}{*}{$\rm Cuboid_2$} & \multirow{2}{*}{Cylinder} & Pentagonal \\
&&&&& prism \\
\hline
ICP \cite{besl1992method} & 0.00/0.00 & 0.00/0.00 & 0.00/0.00 & 0.00/0.00 & 0.00/0.00 \\
ICP w/ kpts \cite{besl1992method} & 0.23/0.22 & 0.55/0.56 & 0.26/0.25 & 0.01/0.00 & 0.08/0.07 \\
3DRHD \cite{nigro2020peg} & 0.02/0.01 & 0.03/0.02 & 0.01/0.00 & 0.00/0.00 & 0.02/0.02 \\
KOVIS \cite{puang2020kovis} & 0.16/0.02 & 0.18/0.03 & 0.26/0.03 & 0.06/0.01 & 0.12/0.02 \\
CFVS (Ours) & \textbf{0.92}/\textbf{0.91} & \textbf{0.98}/\textbf{0.97} & \textbf{0.98}/\textbf{1.00} & \textbf{0.78}/\textbf{0.76} & \textbf{0.92}/\textbf{0.89} \\
\hline
\multicolumn{6}{c}{} \\
\multicolumn{6}{c}{\multirow{2}{*}{(c) 6-DoF peg-in-hole assembly (15 cm / 30 cm initial alignment error)}} \\
\multicolumn{6}{c}{} \\
\hline
\multirow{3}{*}{Method} & \multirow{3}{*}{Avg.} & \multicolumn{2}{c}{Small shape variation} & \multicolumn{2}{c}{Large shape variation} \\
\cline{3-4} \cline{5-6} 
& & \multirow{2}{*}{$\rm Cuboid_1$} & \multirow{2}{*}{$\rm Cuboid_2$} & \multirow{2}{*}{Cylinder} & Pentagonal \\
&&&&& prism \\
\hline
ICP \cite{besl1992method} & 0.00/0.00 & 0.00/0.00 & 0.00/0.00 & 0.00/0.00 & 0.00/0.00 \\
ICP w/ kpts \cite{besl1992method} & 0.21/0.20 & 0.64/0.63 & 0.08/0.07 & 0.05/0.03 & 0.05/0.06 \\
3DRHD \cite{nigro2020peg} & 0.00/0.00 & 0.00/0.00 & 0.00/0.00 & 0.00/0.00 & 0.00/0.00 \\
KOVIS \cite{puang2020kovis} & 0.02/0.02 & 0.04/0.03 & 0.03/0.03 & 0.01/0.00 & 0.00/0.00 \\
CFVS (Ours) & \textbf{0.82}/\textbf{0.82} & \textbf{0.92}/\textbf{0.93} & \textbf{0.90}/\textbf{0.94} & \textbf{0.76}/\textbf{0.72} & \textbf{0.70}/\textbf{0.69} \\
\hline

\end{tabular}
\end{center}
\end{table}

\subsection{Results}
\smallskip\noindent\textbf{Accuracy.} We compare CFVS with four baselines. Table \ref{table:result} depicts the success rate of peg-in-hole assembly with 4 mm clearance. For the 3-DoF task, ICP can hardly complete the insertion task without the accurate information of initial transformation. 3DRHD and ICP w/ kpts have lower performance because they are open-loop methods without further refinement. KOVIS and ours are competitive with 15 cm initial alignment error. However, KOVIS reduces the performance with 30 cm initial alignment error due to the pure visual servoing.

For the 4-DoF task, the success rates of all the baselines are drastically reduced. ICP w/ kpts can still insert into the hole with small shape variation because the target object is similar to the given 3D model. However, they suffer from the large shape variation. KOVIS fails with the square peg and hole. In their experiments, they only test with the round peg and hole, which can not prove that they learn the z-axis rotation. 

For the 6-DoF task, ours has the best performance, showing that CFVS is robust to tilted holes by comprehending the 3D relationship of peg and hole. Moreover, the experiments show that CFVS can be agnostic to objects, solving the problem of large shape variation.

\begin{table}[htbp]
\vspace{6pt}
\caption{Visual servoing time in second. We compare ours with KOVIS \cite{puang2020kovis}. Although KOVIS \cite{puang2020kovis} is faster with 15 cm initial alignment error, they reduce the performance with 30 cm initial alignment error. We keep the same speed with both 15 cm and 30 cm initial alignment errors.}
\label{table:time}
\begin{center}
\vspace{-3mm}
\begin{tabular}{lcc}
\hline
\multirow{2}{*}{Method} & \multicolumn{2}{c}{Initial alignment error} \\
 &15 cm & 30 cm \\
\hline
KOVIS \cite{puang2020kovis} & \textbf{6.6} & 10.1 \\
CFVS (Ours) & 7.0 & \textbf{7.1} \\
\hline

\end{tabular}
\end{center}
\end{table}


\smallskip\noindent\textbf{Efficiency.} In terms of the visual servoing approach, we compare CFVS with KOVIS for the efficiency with 15 cm and 30 cm initial alignment error. For the sake of fairness, we record the visual servoing time in 3-DoF peg-in-hole, as shown in Table \ref{table:time}. We observe that CFVS takes about 7 seconds to complete the task with both 15 cm and 30 cm initial alignment error while KOVIS tends to spend more time with 30 cm initial alignment error. This is because KOVIS needs to move step by step. On the contrary, CFVS achieves the approximate pose at first and then refines the pose to complete the task.

\subsection{Ablations}
\label{sec: ablations}
Table. \ref{table:ablation} demonstrates the result of the ablations. We carry out the ablation study in 6-DoF peg-in-hole with four unseen objects. Our key component is the coarse-to-fine (C2F) framework which boosts the performance a lot. Without refinement, we can not insert into the hole successfully due to the requirement of small error tolerance. Besides, the confidence map (Map) can help the model generalize to unseen objects with the large shape variation. Without the confidence map, we just calculate the average across all $K_i$ and the insertion success rate will decrease. Moreover, data augmentation (Aug) is also a useful strategy that increases the richness of our training set. Without the data augmentation, the performance will be slightly influenced.
\begin{table}[htbp]
\caption{The ablation study. (b,e) shows that the coarse-to-fine approach is crucial to our overall framework. (c,e) shows that using confidence confidence map improves the performance. (d,e) shows that using data augmentation can increase the success rate.
}
\label{table:ablation}
\begin{center}
\setlength{\tabcolsep}{2mm}
\begin{tabular}{cccc|c}
\hline
 & C2F & Map & Aug & Avg. \\
\hline
a & & & & 22\% \\
b & & \Checkmark & \Checkmark & 51\%  \\
c & \Checkmark & & \Checkmark & 73\%  \\
d & \Checkmark & \Checkmark & & 77\%  \\
e & \Checkmark & \Checkmark & \Checkmark & 82\%  \\
\hline

\end{tabular}
\end{center}
\end{table}


\section{Conclusion}
In this paper, we propose CFVS, a coarse-to-fine 3D point-based visual servoing framework, which is the first to achieve the 6-DoF peg-in-hole assembly with tilted holes. CFVS alleviates the issue of large initial error in common visual servoing approaches with a fast and rough pose estimation before gradual refinement. This method greatly reduces the exploration range. In addition, by paying attention to only the information around holes, CFVS generalizes well to unseen objects and is robust to the variation of target shapes. However, although we succeed in 6-DoF, there is still room for improvement. In our 6-DoF experiments, we only test on the square hole and 4 mm clearance. In the future, we can extend our framework to adapt to more different shapes of holes and insert with a tighter clearance.


\section{Acknowledgement}

This work was supported in part by National Science and Technology Council, Taiwan, under Grant NSTC 111-2634-F-002-022, and Mobile Drive Technology Co., Ltd (MobileDrive). We are grateful to the National Center for High-performance Computing.

\bibliographystyle{IEEEtran}
\bibliography{reference}
\end{document}